\documentclass[letterpaper]{article} 
\usepackage[submission]{aaai24}  
\usepackage{times}  
\usepackage{helvet}  
\usepackage{courier}  
\usepackage[hyphens]{url}  
\usepackage{graphicx} 
\usepackage{amssymb} 
\usepackage{makecell}
\usepackage[table]{xcolor}
\usepackage{multirow}
\usepackage{adjustbox}
\usepackage{booktabs}
\usepackage{tabularx}
\usepackage{amsmath}
\usepackage{array}
\usepackage{enumitem}
\usepackage[table]{xcolor}  

\usepackage{natbib}  
\usepackage{caption} 
\usepackage{algorithm}
\usepackage{algorithmic}

\usepackage{newfloat}
\usepackage{listings}
\DeclareCaptionStyle{ruled}{labelfont=normalfont,labelsep=colon,strut=off} 
\floatstyle{ruled}
\newfloat{listing}{tb}{lst}{}
\floatname{listing}{Listing}
\title{\textit{DF-TransFusion}: Multimodal Deepfake Detection via \\ Lip-Audio Cross-Attention and Facial Self-Attention}
\author {
    Aaditya Kharel \textsuperscript{\rm 1},
    Manas Paranjape \textsuperscript{\rm 1},
    Aniket Bera \textsuperscript{\rm 1}
}
\affiliations {
    \textsuperscript{\rm 1}Department of Computer Science, Purdue University, West Lafayette, IN, USA\\
    kharel@purdue.edu, 
    mparanja@purdue.edu, 
    aniketbera@purdue.edu
}

\usepackage{bibentry}

\begin{document}

\maketitle

\begin{abstract}
With the rise in manipulated media, deepfake detection has become an imperative task for preserving the authenticity of digital content. In this paper, we present a novel multi-modal audio-video framework designed to concurrently process audio and video inputs for deepfake detection tasks. Our model capitalizes on lip synchronization with input audio through a cross-attention mechanism while extracting visual cues via a fine-tuned VGG-16 network. Subsequently, a transformer encoder network is employed to perform facial self-attention. We conduct multiple ablation studies highlighting different strengths of our approach. Our multi-modal methodology outperforms state-of-the-art multi-modal deepfake detection techniques in terms of F-1 and per-video AUC scores.
\end{abstract}

\label{sec:intro}
\section{Introduction}

Deepfake refers to the application of deep learning techniques for generating manipulated digital media, such as video or audio, often with malicious intent for purposes like fraud, defamation, and the dissemination of disinformation or propaganda. The proliferation of deepfakes poses a significant threat to the authenticity and credibility of digital media, causing adverse effects on businesses, governments, and political leaders as it becomes increasingly difficult for humans to discern whether a piece of audio or video has been manipulated.

Although multimedia forgery is not a novel phenomenon, advancements in deep learning and the development of generative adversarial networks (GAN) \cite{goodfellow2020generative, 8253599, 8039016} have revolutionized the field of computer vision and deepfake generation. State-of-the-art techniques, such as FaceSwap, FakeApp, FaceShifter \cite{li2020advancing}, Face2Face \cite{10.1145/3292039}, DeepFaceLab \cite{https://doi.org/10.48550/arxiv.2005.05535}, and Neural Textures \cite{thies2019deferred}, have been employed to create deepfakes by swapping faces in original videos with target images. The widespread availability of deepfake generation methods necessitates the development of sophisticated deep learning techniques to combat the issue.

\begin{figure}[h!]
    \centering
    \includegraphics[width=0.45\textwidth]{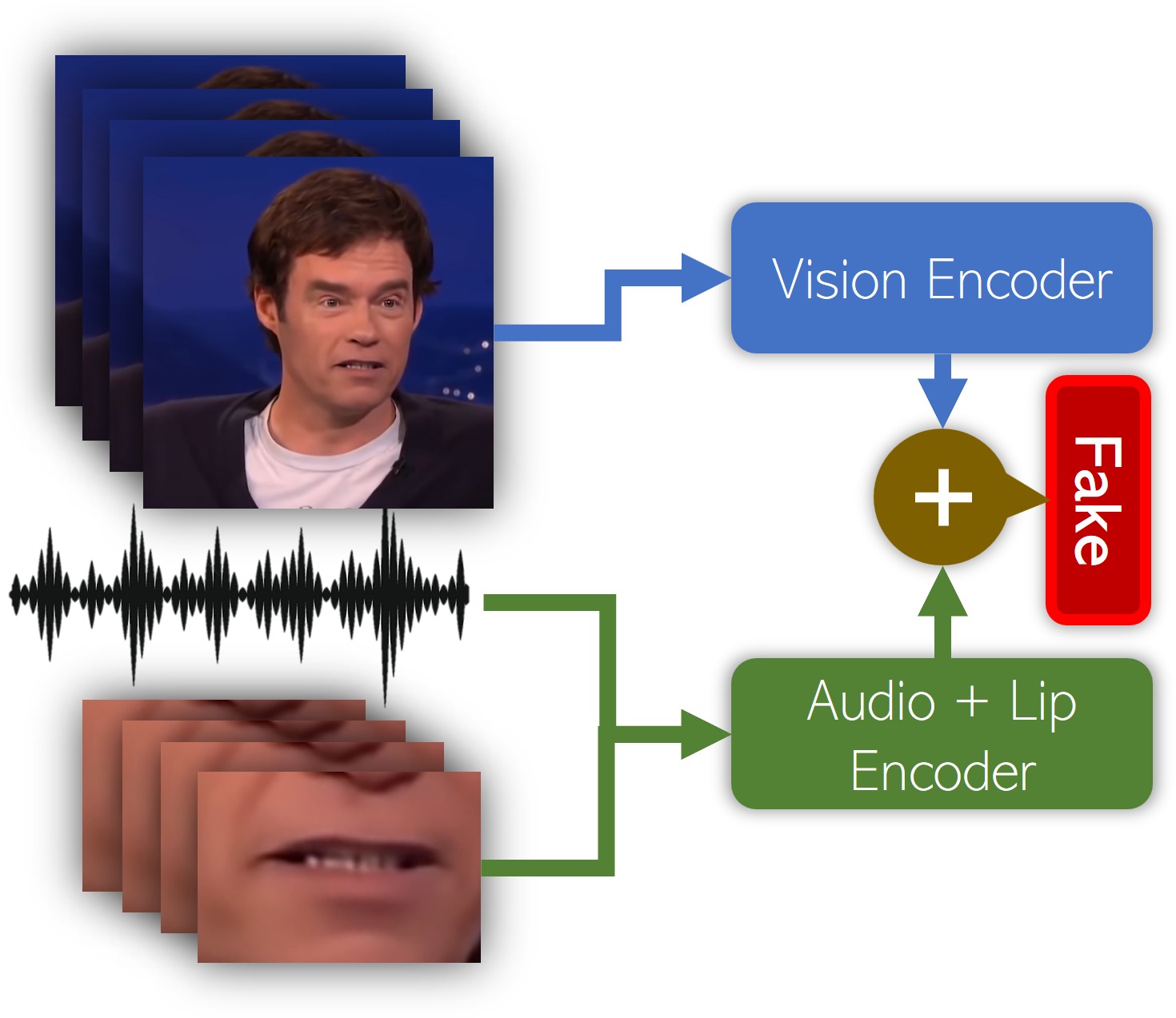}
    \caption{We propose a multi-modal deepfake detection technique that uses self-attention to detect deepfake artifacts and cross-attention to identify discrepancies between lip movements and audio signals.}
    \label{fig:cover}
\end{figure}

Several convolutional neural network (CNN) architectures have been proposed for deepfake detection \cite{zhou2017two, afchar2018mesonet, li2018exposing, faceforensics++, MultiTask, yang2018learning}. Recurrent neural networks have also been utilized to capture time dependencies in deepfake detection tasks \cite{guera2018deepfake, masi2020two}. More recently, transformer-based architectures with multi-head attention mechanisms \cite{vaswani2017attention, bdcc5040049, zhao2021multi} have demonstrated promising results compared to CNN-based methods. Some approaches even combine CNN and attention mechanisms for deepfake detection \cite{coccomini2022combining}. However, most deepfake detection techniques focus on either video or audio modalities, with only a few addressing both \cite{mittal2020emotions}. Given the rise in accessibility of deepfake generation, it is vital to develop deepfake detection methods that consider both audio and video modalities.

The majority of deepfake detection models focus on either audio-only or video-only detection methods, primarily due to the scarcity of datasets featuring both audio and video deepfakes. Datasets like UADFV \cite{yang2019exposing}, FaceForensics++ \cite{faceforensics++}, CelebDF \cite{li2019new}, and Deeper Forensics 1.0 \cite{jiang2020deeperforensics} contain video-only deepfakes. In contrast, DFDC \cite{dolhansky2020deepfake}, DF-TIMIT \cite{korshunov2018deepfakes}, and FakeAVCeleb \cite{khalid2021fakeavceleb} feature deepfakes in both audio and video modalities. Ignoring the audio modality can be problematic, as audio provides crucial information for multimodal deepfake detection tasks \cite{mittal2020emotions}. Table ~\ref{subsec:datasets} summarizes deepfake dataset based on the modality in which the deepfake occurs.

In this paper, we address the limitations of existing deepfake detection approaches by proposing a multi-modal method that effectively leverages both audio and video information for deepfake detection. We introduce a novel pipeline that employs a fine-tuned VGG-16 feature extractor and transformer encoders to process and analyze input audio and video data. Our approach utilizes self-attention mechanisms to detect deepfake artifacts in facial regions, and cross-attention mechanisms to identify discrepancies between lip movements and audio. We evaluate the performance of our method through rigorous ablation studies and demonstrate that our multi-modal approach outperforms state-of-the-art methods in deepfake detection, even when compared to unimodal approaches with significantly more trainable parameters. To summarize, we propose the following: 

\begin{itemize}
\item We introduce a novel multi-modal deepfake detection technique that capitalizes on both audio and video modalities by employing fine-tuned VGG-16 feature extractor and transformer encoders.

\item Our method uses self-attention to detect deepfake artifacts in facial regions, and cross-attention to identify discrepancies between lip movements and audio signals. 

\item Our proposed approach surpasses the previous state-of-the-art in multi-modal deepfake detection strategies

\item We conduct comprehensive ablation studies to demonstrate the efficacy of our approach. 
\end{itemize}

In Section \ref{sec:relatedwork}, we perform a thorough literature review on media forensics as well as unimodal and multimodal deepfake detection methods. In Section \ref{sec:approach}, we discuss our model architecture and data pre-processing in detail. In Section \ref{sec:exp}, we show our experimental results on multimodal baseline methods as well as results from our ablation study. In Section \ref{sec:conclusion}, we provide concluding remarks. Finally, in section \ref{sec:future work}, we provide direction for future work in deepfake detection tasks. 

\section{Related Work}
\label{sec:relatedwork}
In this section, we provide an overview of previous work in the domain. First, we discuss multimedia forensics literature. Second, we review unimodal deepfake detection methods. Third, we examine multimodal approaches for deepfake detection. 

\begin{figure*}[h!]
\begin{center}
\includegraphics[width=\linewidth]{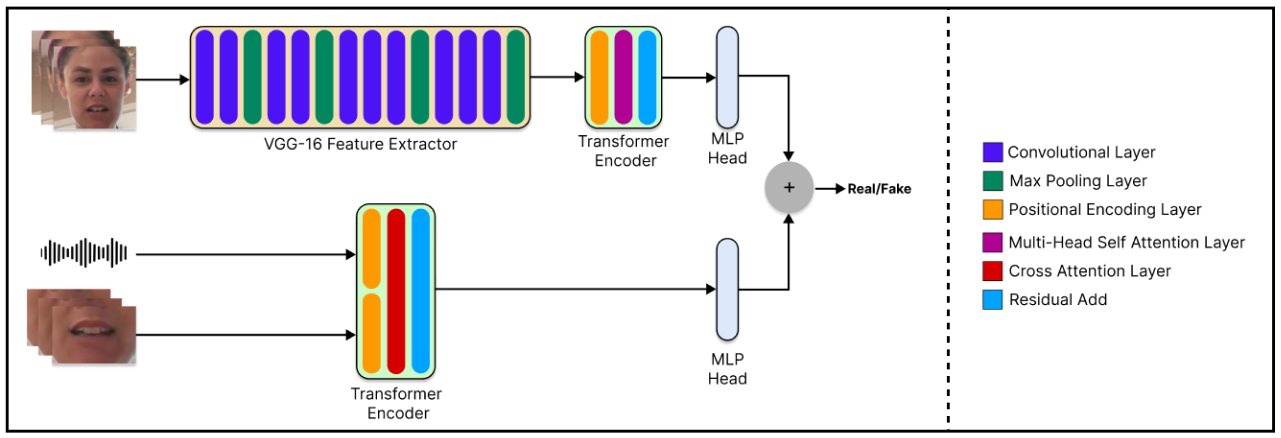}
\end{center}
   \caption{Our model architecture shows input video frames, audio and lip region along with fine-tuned VGG-16 feature extractor, transformer encoders, and multi-layer perceptron heads.}
\label{fig:arch}
\end{figure*}
\subsection{Media Forensics}
\label{subsec:forensics}
Multimedia forensics tackles verification and authenticity of digital media sources to detect forgeries and malicious contents~\cite{for1}. Traditional computer-vision methods~\cite{for1,for4} have been adequate for dealing with multimedia forgery, which was not generated using deep learning methods. However, almost all deepfakes that are recently created use deep learning to tamper with the media. For example, ~\cite{DeepFakesMalicious} leverages artificial intelligence to generate highly realistic and persuasive fake media. Identifying false media that has been manipulated with deep learning proves challenging for classical computer vision techniques~\cite{for7}. As a result, there is a growing interest in creating deep learning-based solutions for multimedia forensics~\cite{for3,for5,for6}. Prior studies have also investigated the role of affect in deception detection~\cite{liar}.

\begin{table}[t]
\addtolength{\tabcolsep}{-3pt}
\begin{tabular}{|l|l|l|l|}
\hline
\textbf{Dataset} & \textbf{\# Videos} & \textbf{Visual} & \textbf{Audio} \\
\hline
\makecell[l]{UADFV \\ \cite{yang2019exposing}} & 98 & \checkmark & $\times$ \\
\hline
\makecell[l]{DF-TIMIT \\ \cite{korshunov2018deepfakes}} & 320 & \checkmark & \checkmark \\
\hline
\makecell[l]{Face Forensics++ \\ \cite{faceforensics++}} & 5,000 & \checkmark & $\times$ \\
\hline
\makecell[l]{CelebDF \\ \cite{Celeb-DF}} & 1,203 & \checkmark & $\times$ \\
\hline
\makecell[l]{DFDC \\ \cite{dolhansky2020deepfake}} & 119,146 & \checkmark & \checkmark \\
\hline
\makecell[l]{Deeper Forensics 1.0 \\ \cite{jiang2020deeperforensics}} & 60,000 & \checkmark & -- \\
\hline
\makecell[l]{FakeAVCeleb \\ \cite{khalid2021fakeavceleb}} & 21,566 & \checkmark & \checkmark \\
\hline
\end{tabular}
\caption{Table showing various deepfake datasets along with the modality in which deepfake occurs. There are very few datasets where deepfake occurs on both audio and video modality, hence making multimodal experimentation limited.}
\label{subsec:datasets}
\end{table}

\subsection{Unimodal DeepFake Detection Methods}
\label{subsec:unimodal}
Most previous research in deepfake detection focuses on dissecting videos into individual frames and analyzing the visual discrepancies within them. For example, \cite{li2018exposing} suggest a Deep Neural Network (DNN) for detecting fake videos by examining artifacts present during the face distortion phase of generation methods. In a similar vein, \cite{yang2019exposing} explore inconsistencies in head positions in synthesized videos, while \cite{VA-MLP} identifies anomalies in the eyes, teeth, and facial outlines of generated faces. Prior work has also experimented with a variety of network architectures, such as \cite{MultiTask} investigating capsule structures, \cite{faceforensics++} utilizing XceptionNet, and \cite{zhou2017two} adopting a two-stream Convolutional Neural Network (CNN) to attain state-of-the-art performance in general-purpose image forgery detection. Additionally, researchers have noticed and taken advantage of the fact that temporal coherence is not effectively maintained during deepfake synthesis. For instance, \cite{temporal1} employs spatio-temporal characteristics of video sequences to detect deepfakes, and \cite{guera2018deepfake} highlights the presence of intra-frame consistencies in deepfake videos, leading them to implement a CNN combined with a Long Short Term Memory (LSTM) for deepfake video identification.

There are multiple audio deepfake detection techniques as well. Many prior methods that have evaluated on ASVspoof2021 dataset have used the model ensemble technique. For example, \cite{mo2020automatic} propose using two feature extractions method, namely, Mel-Frequency Cepstral Coefficients (MFCC) and Constant Q Cepstral Coefficients (CQCC), to evaluate the prediction accuracy using SVM and Gaussian Mixture Model (GMM). Similarly, \cite{9746163} proposes a two-path spoofing detection method where authors use Linear Frequency Cepstral Coefficients (LFCC) and CQCC to extract the features. Then, one of the paths contains real-GMM and the other path contains fake-GMM. The output from each GMM, which contains the Gaussian probability feature, is then fed into their respective identical Res \cite{he2016deep} blocks. The output from two Conv networks is then concatenated to feed into a fully connected layer for output. As part of an automated end-to-end pipeline, \cite{wang2022fully} proposed Wav2Vec for feature extraction and representation for unlabelled speech data. 


\subsection{Multimodal DeepFake Detection Methods}
\label{subsec:multimodal}
Although unimodal DeepFake Detection approaches (mentioned in Section~\ref{subsec:unimodal}) have predominantly concentrated on an individual's facial characteristics, the inclusion of multiple modalities within the same video has received limited attention. \cite{multimodal1} introduce FakeTalkerDetect, which is a Siamese-based network devised for identifying fake videos produced by neural talking head models. The use of lip-sync mechanisms for deepfake detection is not a novel idea. For instance, Haliassos \cite{haliassos2021lips} have already used lip-synchronization using ResNet-18 as a feature extractor and temporal convolutional network (MS-TCN) on DFDC. However, the reported AUC is inferior compared to our model's AUC score. Other various works on lip-sync exist but not necessarily for the downstream task of deepfake detection. For instance,  \cite{halperin2019dynamic} have used the temporal alignment of the lip-to-speech. \cite{li2018exposing} leverage face-warping artifacts for deepfake detection. Likewise, \cite{gu2021spatiotemporal} uses spatio-temporal inconsistencies for deepfake detection tasks. Similarly, \cite{chugh2020not, haliassos2021lips} uses disharmony between the audio and video modality to detect lip-sync inconsistencies. Similarly, \cite{zhou2021joint} suggests that deepfake detection can benefit a lot from synchronizing the audio modality with the video modality. \cite{haliassos2021lips} focuses entirely on lipreading mechanisms to detect deepfake. However, if deepfake does not occur in the mouth region, then the model fails to detect deepfake. Our model addresses this issue by not only performing lip-audio cross-attention but also facial self-attention, encapsulating eclectic deepfake scenarios.

\section{Our Approach}
\label{sec:approach}


Our proposed method employs a fine-tuned VGG-16 feature extractor in conjunction with transformer encoder modules to address deepfake detection tasks. We begin by extracting the facial region from the video using the MTCNN network. Simultaneously, we feed the raw audio and the extracted lip region into the audio transformer encoder module. To process the data, facial self-attention is applied for the video-only pipeline, while cross-attention is employed for lip-audio synchronization analysis.

The outputs from both the audio and video transformer encoder modules are subsequently passed through a multi-layer perceptron (MLP) head, which generates the final classification label, as illustrated in Figure \ref{fig:arch}. In the following sections, we provide a comprehensive explanation of our approach, encompassing the pre-processing steps involved. Table \ref{tab:variables} lists the variables and their dimensions used in our approach.

\begin{table}[h!]
\centering
\addtolength{\tabcolsep}{-3pt}
\begin{tabular}{|p{0.06\textwidth}|l|}
\hline
\textbf{Var} & \textbf{Dimensions} \\\hline
$N_f$ & Number of temporal divisions per video \\\hline
$N_p$ & Number of spatial divisions per frame \\\hline
$P$ & Size of each spatial division \\\hline
$F$ & Size of each temporal division \\\hline
$\textbf{x}$ & Feature extracted video patches \\\hline
$\textbf{x}p$ & Flattened video patches \\\hline
$\textbf{E}$ & Linear projection layer \\\hline
$\textbf{E}{pos}$ & Positional embedding \\\hline
$\textbf{z}_0$ & Initial patch embedding \\\hline
$\textbf{z}^{'}_1$ & Output of multi-head self-attention layer \\\hline
$\textbf{z}_1$ & Output of MLP \\\hline
$\textbf{y}$ & Final video representation \\\hline
$d$ & Dimension of the transformer embedding output \\\hline
$\textbf{y}_a$ & Output embedding for audio transformer\\\hline
$\textbf{y}_v$ & Output embedding for video transformer\\\hline
\end{tabular}
\caption{Variables and their dimensions used in our approach}
\label{tab:variables}
\end{table}

\subsection{Video Preprocessing}

Our method for detecting deepfakes in videos primarily focuses on the facial region, as this is where most deepfake manipulations occur. We first use the MTCNN network to extract facial bounding boxes throughout the video. From these bounding boxes, we crop the facial regions and isolate them for further analysis.

\subsection{Video Frame Rates}
\subsubsection{Video Self-Attention}
The DFDC dataset's video frame rate is approximately $30$ frames per second, resulting in $300$ frames per video. However, computing attention across all $300$ frames for each video is computationally expensive. Since most deepfakes in the dataset are present throughout the entire video, we can sample equally spaced frames for the video section of the transformer. Therefore, we use only $30$ equally spaced frames in the video-only pipeline.

\subsubsection{Audio Cross-Attention}
To perform lip-syncing, we require all $300$ frames to correlate with the audio. We extract lip regions from the faces and use all $300$ frames for cross-attention to perform lip-syncing. These images are also converted to low-resolution, black-and-white images, as color is irrelevant for lip-syncing purposes.

\subsection{Image Preprocessing}
\subsubsection{Video Self-Attention}
After extracting 30 frames from the video, we preprocess these images by resizing them to a final size of $256\times256\times3$, where 3 is the number of channels $(R, G, B)$. We feed each of these images individually to our fine-tuned VGG-16 feature extractor.

\subsubsection{Audio Cross-Attention}
During the extraction of the 300 frames from the video, we crop the frames to get only the lip regions. We crop these images as well to get a final image size of $35 \times 140$. We feed these images as one of the inputs to the audio cross-attention layer of the audio transformer encoder block.

\subsection{Audio Preprocessing}
The input audio for all videos in the dataset is recorded at a standard rate of $44.1KHz$. We first convert this audio from stereo to mono to reduce the input size by half without affecting lip-sync functionality. We crop the audio to the first $441,000$ values, as all videos in the dataset are almost exactly 10 seconds long. This does not result in significant data loss, as the average number of values in the input tensor is $441,300$. This audio is then fed into the audio cross-attention layer of the audio transformer encoder block.

\begin{figure}[t]
    \centering
    \includegraphics[width=5cm]{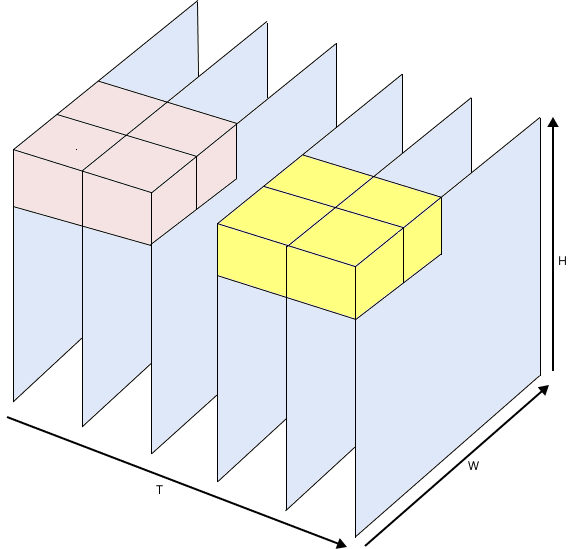}
    \caption{Cuboid Embedding for spatio-temporal 3-D Attention}
    \label{fig:tubelet}
\end{figure}

\subsection{Cuboid Embedding} 
We use tubelet embedding, as suggested in \cite{arnab2021vivit}, to capture the spatio-temporal dimension of the input video frames. A tubelet is a 3-D volume that captures the height ($h$), width ($w$), and depth ($t$) of the frames. The total number of tokens extracted in each dimension can be given as:
\begin{align*}
    &\text{Number of tokens in time} (n_t) = \Big \lfloor \frac{T}{t} \Big \rfloor \\
    &\text{Number of tokens in 
    frame height} (n_h) = \Big \lfloor \frac{H}{h} \Big \rfloor \\
    &\text{Number of tokens in frame width} (n_w) = \Big \lfloor \frac{W}{w} \Big \rfloor
\end{align*}
where, $H, W, T$ correspond to the frame height $(H)$, frame width $(W)$ and depth or number of frames in temporal dimension $(T)$. The tubelet embedding has been shown in Fig \ref{fig:tubelet}. The tubelet embedding method captures both spatial and temporal relationships between frames simultaneously, unlike the uniform frame sampling approach, where each 2D frame is tokenized independently, and temporal embedding information must be provided separately to the transformer encoder. Using tubelet embedding simplifies computing cross-attention between the lip region and audio as temporal synchronization is not required since the tubelet embedding inherently preserves both positional and temporal information.


\subsection{Self-Attention Mechanism}

For the video-only pipeline, we use multi-head self-attention to compute different attention filters. This is achieved by using a transformer encoder with multiple sets of query ($Q$), key ($K$), and value ($V$) inputs to obtain $n$ attention layers. Self-attention means that the initial query, key, and value are all equal (i.e., $Q=K=V$) and come from the image patches of the detected faces. Our processing pipeline is similar to that used in Vision Transformer (ViT) \cite{https://doi.org/10.48550/arxiv.2010.11929} and Convolutional Cross EfficientNet \cite{coccomini2022combining}.

In the video pipeline, attention is computed by first computing the softmax of the scaled cosine similarity between $Q$ and $K$, and then multiplying it with $V$ to compute the attention filter, as shown in Equation \ref{eq:attention} below:

\begin{align}
\label{eq:attention}
Attention(Q,K,V) = softmax\Big(\frac{QK^{T}}{\sqrt{d_k}}\Big)V
\end{align}

In the self-attention mechanism used in the video pipeline, $Q = K = V \in \mathbb{R}^{3073 \times 1280}$. This approach allows us to capture the spatio-temporal relationships in the video and compute cross-attention between the lip region and the audio without needing to synchronize them in time.

 \subsection{Video Transformer}
Figure \ref{fig:arch} illustrates how we process input for the video branch in the transformer encoder. In this section, we define the important parameters and their dimensions. The output of the VGG-16 feature extractor, which is the transformer encoder input, consists of $N_p$ patches of size $P \times P$ for each of the $N_f$ frames. We denote the output of the feature extractor as $\textbf{x} \in \mathbb{R}^{N_f \times N_p \times {P^2}}$. We partition these $N_f$ patches into $\frac{N_f}{F}$ parts, each with size $F$, and then combine each of these parts to get 3-dimensional patches of size $P^2 \cdot F$, which are then flattened. This can be written as:

\begin{equation}
\textbf{x}_p \in \mathbb{R}^{\frac{(N_p \cdot N_f)}{F} \times (P^2 \cdot F)}
\end{equation}

Once the frames are rearranged in this format, we pass them through a linear layer $\textbf{E}$ of dimension $(P^2 \cdot F ) \times (3 \cdot P^2 \cdot F )$ similar to \cite{https://doi.org/10.48550/arxiv.2010.11929}. The output of this projection is referred to as the patch embedding. The process is shown as:

\begin{align}
    \textbf{z}_0 &= [\textbf{x}_{CLS}; \textbf{x}^{i}_{p}\textbf{E}] + \textbf{E}_{pos} \\
    \mathbf{z^{'}}_1 &= MSA(LN(\mathbf{z}_{0})) + \mathbf{z}_{0}\\
    \mathbf{z}_1 &= MLP(LN(\mathbf{z^{'}}_{1})) + \mathbf{z}_{1}\\
     \textbf{y} &= LN(z_1)
\end{align}

Here, $x^{i}p$ refers to the flattened $i^{th}$ video patches, and $\textbf{x}_{CLS}$ is the prepended classification token. $\textbf{E}{pos}$ is the positional embedding, and $LN$ is the layer normalization. $MSA$ is the multi-headed self-attention, and $MLP$ is the multi-layer perceptron, as in \cite{devlin2018bert, dosovitskiy2020image}. The prepended classification token ($\textbf{z}_{0}^{0}=\textbf{x}_{CLS}$) is a learnable embedding whose state at the output of the Transformer encoder ($\textbf{z}_{1}^{0}$) serves as the video representation $\textbf{y}$. The positional embedding $\textbf{E}_{pos}$ helps maintain the positional information.

\subsection{Cross-Attention Mechanism}
To enable cross-attention between the video and audio modalities for lip-synchronization and lip-audio consistency, we need to consider the different dimensions of $Q$, $K$, and $V$. Specifically, $Q$ $\in \mathbb{R}^{300 \times 4900}$ from the video-only pipeline, while $K$ and $V \in \mathbb{R}^{300 \times 1470}$ from the lip-audio pipeline. Since $Q$ and $K$ need to have the same dimensions for computing the cosine similarity, as shown in Equation \ref{eq:attention}, we pass $Q$ through a linear layer $L_1 \in \mathbb{R}^{4900 \times 4900}$, which reshapes the matrix $Q$ into $Q^{\prime} \in \mathbb{R}^{300 \times 4900}$. Similarly, we reshape $K$ using a linear layer $L_2 \in \mathbb{R}^{1470 \times 4900}$ into $K^{\prime} \in \mathbb{R}^{300 \times 4900}$ to ensure the compatibility of dimensions. This is shown below:
\begin{align}
Q \cdot L_1 &= Q^{\prime}, & Q^{\prime} \in \mathbb{R}^{300\times 4900}, \\
K \cdot L_2 &= K^{\prime}, & K^{\prime} \in \mathbb{R}^{300\times 4900}
\end{align}
After this reshape, we compute the attention filter or the cosine similarity between $Q^{\prime}$ and $K^{\prime}$, pass it through a softmax and then finally multiply it with $V$. That is, we compute the similarity in the same way as in Equation \ref{eq:attention}, except that $Q=Q^{\prime}$ and $K=K^{\prime}$.

\subsection{Multi-Layer Perceptron (MLP)}
In the final stage of our approach, we pass the output embeddings from the transformer encoder modules of both the video and audio modalities through a multi-layer perceptron (MLP) head to obtain the final classification label. Let $\textbf{y}_a$ and $\textbf{y}_v$ denote the output embeddings from the audio and video transformer encoder blocks, respectively. We concatenate these two embeddings to obtain a joint embedding $\textbf{y} \in \mathbb{R}^{2d}$, where $d$ is the dimension of the transformer embedding output, and in our case, $d=2$. We then pass this joint embedding through a linear layer of size $d\times2$ to output the final probability for real and fake classification.

\section{Experiments and Results}
\label{sec:exp}
In this section, we describe our experimental setup as well as demonstrate the quantitative and qualitative results of our model.

\begin{table}[h!]
\centering
\renewcommand{\arraystretch}{1.1}
\begin{tabular}{|m{0.6\linewidth}|>{\centering\arraybackslash}m{0.12\linewidth}|>{\centering\arraybackslash}m{0.12\linewidth}|}
\hline
\textbf{Model} & \textbf{DFDC AUC} & \textbf{DF-TIMIT AUC} \\
\hline
\rowcolor{gray!20}\multicolumn{3}{|c|}{\textbf{Multimodal Baseline Approaches}} \\
\hline
Capsule~\cite{Capsule} & 0.533 & 0.744 \\
\hline
Multi-task~\cite{MultiTask} & 0.536 & 0.553 \\
\hline
HeadPose~\cite{yang2019exposing} & 0.559 & 0.532 \\
\hline
Two-stream~\cite{zhou2017two} & 0.614 & 0.735 \\
\hline
VA-MLP~\cite{VA-MLP} & 0.619 & 0.621 \\
\hline
VA-MLP~\cite{VA-MLP} & 0.662 & 0.773 \\
\hline
MesoInception4~\cite{afchar2018mesonet} & 0.732 & 0.627 \\
\hline
Meso4~\cite{afchar2018mesonet} & 0.753 & 0.684 \\
\hline
Xception-raw~\cite{faceforensics++} & 0.499 & 0.540 \\
\hline
Xception-c40~\cite{faceforensics++} & 0.697 & 0.705 \\
\hline
Xception-c23~\cite{faceforensics++} & 0.722 & 0.944 \\
\hline
FWA~\cite{li2018exposing} & 0.727 & 0.932 \\
\hline
DSP-FWA~\cite{li2018exposing} & 0.755 & 0.997 \\
\hline
Emotions Don't Lie~\cite{mittal2020emotions} & 0.844 & 0.949 \\
\hline
\rowcolor{blue!20}\textbf{Our Approach} & \textbf{0.979} & \textbf{1.000} \\
\hline
\end{tabular}
\caption{Comparison of AUC scores on DFDC and DF-TIMIT with the prior baseline methods. Our multimodal method beats prior state-of-the-art in both DFDC and DF-TIMIT. Note that prior methods in this table have not reported the F-1 score.}
\label{tab:baselinemm}
\end{table}


\subsection{Experimental Setup}
We tested our model on the DFDC dataset, DF-TIMIT and FakeAVCeleb, which are the only known dataset containing deepfake in both audio and video modality. We leveraged four A100 (80 GB) GPUs with NVLink for training our model. Each GPU used approximately 40 GB of memory during the training. The batch size is 8 videos with a total of 1310 batches per epoch. The loss function used is the Cross-Entropy loss with a learning rate of $10^{-6}$. When computing the accuracy for each epoch, we compute the accuracy for each batch and average the accuracy across all batches in every epoch to report accuracy per epoch. We show that our method strongly outperforms all existing methods on multimodal deepfake detection in Table ~\ref{tab:baselinemm}.


\newcolumntype{C}[1]{>{\centering\arraybackslash}m{#1}}

\subsection{Qualitative Results}
\label{subsec:qual}
\begin{figure*}[h!]
\begin{center}
\includegraphics[width=1\linewidth]{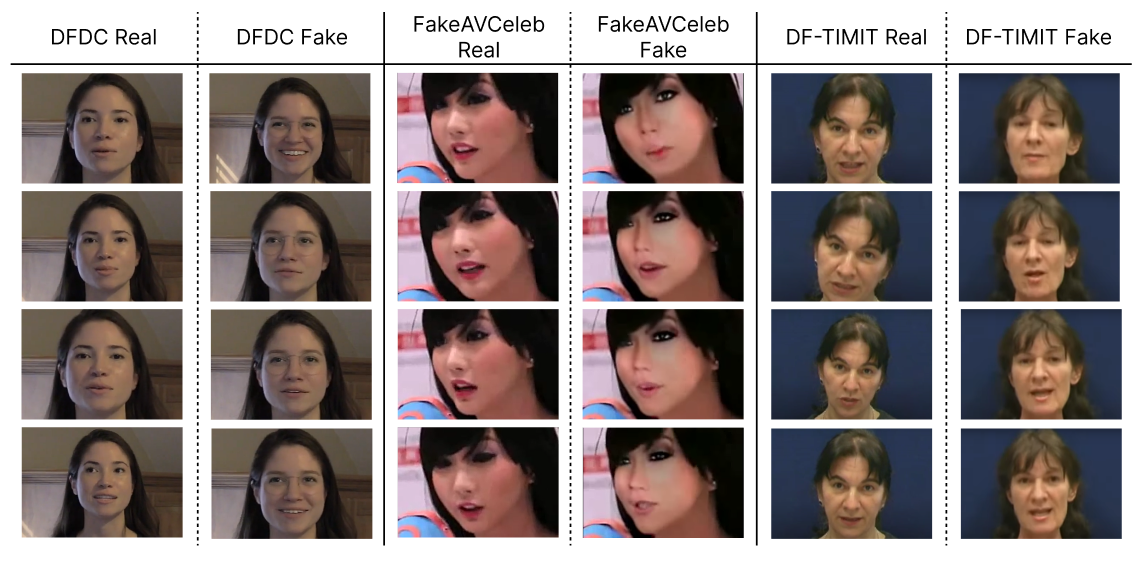}
\end{center}
   \caption{Qualitative Results: We show some sample frames from the DFDC, FakeAVCeleb, and DF-TIMIT that the model uses during training and testing. Our model uses both audio-video modality and the cross-attention between the two modalities  to classify as real and fake videos.}
\label{fig:dfdc}
\end{figure*}

In Figure ~\ref{fig:dfdc}, we display some selected frames of videos from the DFDC, DF-TIMIT, and FakeAVCeleb datasets, along with the corresponding labels (real or fake). Note that the real videos in the DF-TIMIT section of Figure ~\ref{fig:dfdc}  are not in the testing set, but the original videos from which the deepfakes were generated. Overall, our qualitative results demonstrate the effectiveness of our proposed approach in detecting deepfaked videos, particularly in capturing abnormal features in such videos.

\subsection{Ablation Study}
To evaluate the contribution of different components of our proposed network architecture, we conduct two ablation studies. 

\subsubsection{Lip-Audio Only}
The ablated pipeline utilized in our model involved both audio and video modalities but only focused on lip-sync between the audio and lip region of the video. While this model learned to differentiate between real and fake videos to an extent, solely relying on lip-audio sync is not enough, as evidenced by ~\ref{tab:baseline2}. As a result, we maintain that the video pipeline is integral to getting better results due to the lack of visual features, which are omitted for the audio-lip pipeline.

\subsubsection{Effect of VGG-16 (Removing Transformer Encoder)}
Likewise, we also tried removing the transformer encoder from our unimodal pipeline and performed the classification based on the features extracted by our fine-tuned VGG-16 architecture. Our results show that the F-1 score of such a fine-tuned VGG-16 architecture is 87.77\% with and AUC score of 0.944 as shown in Table ~\ref{tab:baseline2}. While the F1-score is high for an independent VGG-16 network, the addition of transformers after feature extraction still provides valuable gains.


\begin{table}[t]
\centering
\renewcommand{\arraystretch}{1.2}
\begin{tabular}{|c|c|c|}
\hline
\textbf{Model} & \textbf{AUC} & \textbf{F1} \\
\hline
\rowcolor{gray!20}\multicolumn{3}{|c|}{\textbf{Performance Evaluation}} \\
\hline
DFDC & 0.979 & 92.7 \\
\hline
FakeAVCeleb & 0.748 & 84.8 \\
\hline
DF-TIMIT & 1.000 & 100.0 \\
\hline
\end{tabular}
  \caption{Performance Evaluation on 3 different datasets}
  \label{tab:methodsummary}
\end{table}


\begin{figure*}[h!]
\begin{center}
\includegraphics[width=1\linewidth]{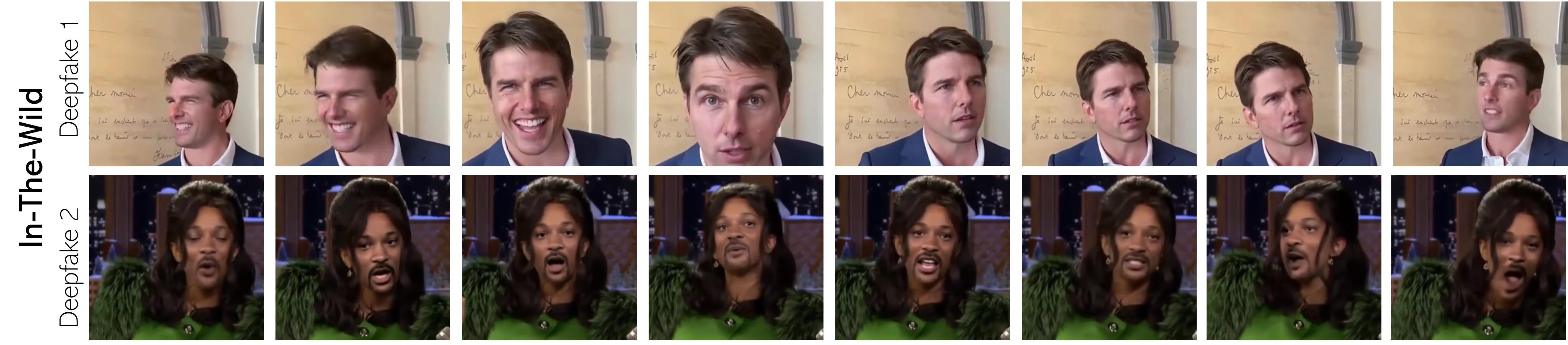}
\end{center}
   \caption{Evaluation on In-The-Wild Videos: To evaluate the robustness of our model, we tested it on out-of-dataset videos. Our model correctly detected sample videos downloaded from Youtube, MIT Deepfake Lab, and other public sources. In this image, we show that our method was able to identify a recently popular DeepFake video of celebrities Tom Cruise and Will Smith.}
\label{fig:wild}
\end{figure*}


\begin{table}[t]
\centering
\renewcommand{\arraystretch}{1.2}
\begin{tabular}{|p{0.38\linewidth}|c|c|c|}
\hline
\textbf{Model} & \textbf{AUC} & \textbf{F1} & \textbf{\# param} \\
\hline
\rowcolor{gray!20}\multicolumn{4}{|c|}{\textbf{Ablation Study}} \\
\hline
Multimodal Approach & 0.979 & 92.7 & 90M \\
\hline
Lip+Audio Only & 0.452 & 63.7 & 62M \\
\hline
Fine-Tuned VGG-16 Features Only & 0.944 & 87.7 & 7M \\
\hline
\end{tabular}
\caption{Ablation Study: Comparison of AUC and F1 scores evaluation metric on DFDC dataset after removing parts of the multimodal network.}
\label{tab:baseline2}
\end{table}

\subsection{Failure Cases}
\label{subsec:failure}
While our deepfake detection approach achieves high accuracy on all multimodal datasets, it still has some failure cases. One of the most common causes of failure is when the face is not directly pointing toward the camera. In such cases, our approach may not be able to capture enough features of the face and lip movements, which can result in inaccurate predictions.

Another issue that can lead to failure is when the video is blurry or has low resolution. Similarly, when two speakers are facing each other, our approach may detect the non-speakers face, leading to incorrect predictions. This can happen when the non-speakers face is more visible or when the speaker's lips and facial region are not clearly visible.

\subsection{Results on Videos in the Wild}
\label{subsec:wild}

To evaluate the effectiveness of our deepfake detection model, we conducted experiments on multiple deepfake videos obtained from various online platforms, specifically YouTube and MIT Deepfake Lab. We visually analyzed selected frames from these videos, as shown in Figure~\ref{fig:wild}, and applied our model to detect whether the videos were real or fake. The results of our experiments showed that our model was successful in correctly classifying most of the deepfake videos.

\section{Conclusion}
\label{sec:conclusion} 
In this work, we proposed a novel deepfake detection method that leverages both audio and visual modalities. Our proposed approach achieved state-of-the-art performance on the DFDC and DF-TIMIT, which are all multimodal deepfake detection datasets, outperforming eleven prior deepfake detection methods. We also achieve close to state-of-the-art performance on the FakeAVCeleb dataset independent of deepfake generation methodology. We also provide ablations on sections of the model that perform less than optimally compared to the multimodal approach. Additionally, we evaluated our approach on in-the-wild videos and demonstrated promising results.

\section{Future Work}
\label{sec:future work}
One primary issue with deepfake datasets is that there is a major class imbalance between the real and the fake samples, which becomes a bottom-neck when creating an unbiased model. Either larger and more balanced datasets are necessary, or methods that can handle class imbalance are necessary. Current deepfake detection methods cannot adapt to multi-speaker situations in videos which is another promising future work direction. Model performance is also impacted if data is noisy due to factors, such as poor lighting conditions and camera angle, i.e., if participants are not directly facing the camera. Future deepfake detection requires techniques that are beyond facial and audio analysis. 

\bibliography{aaai24}

\end{document}